\documentclass[conference]{IEEEconf}
\pdfoutput=1
\IEEEoverridecommandlockouts

\usepackage{cite}
\usepackage{amsmath,amssymb,amsfonts}
\usepackage{algorithmic}
\usepackage{graphicx}
\usepackage{textcomp}
\usepackage{xcolor}
\usepackage{bm}
\def\BibTeX{{\rm B\kern-.05em{\sc i\kern-.025em b}\kern-.08em
    T\kern-.1667em\lower.7ex\hbox{E}\kern-.125emX}}
\begin{document}

\title{HumanFT: A Human-like Fingertip Multimodal Visuo-Tactile Sensor}

\author{Yifan Wu, Yuzhou Chen$^+$, Zhengying Zhu$^+$, Xuhao Qin$^+$, and Chenxi Xiao$^*$
\thanks{$^+$equal contribution}
\thanks{*This work was supported by Shanghai Frontiers Science Center of Human-centered Artificial Intelligence (ShangHAI), MoE Key Laboratory of Intelligent Perception and Human-Machine Collaboration (KLIP-HuMaCo). The experiments of this work were supported by the Core Facility Platform of Computer Science and Communication, SIST, ShanghaiTech University.}
\thanks{Yifan Wu is with the School of Information Science and Technology at ShanghaiTech University,  {\tt\footnotesize wuyf2024@shanghaitech.edu.cn}}%
\thanks{Yuzhou Chen is with the School of Information Science and Technology at ShanghaiTech University,  {\tt\footnotesize chenyzh3@shanghaitech.edu.cn}}%
\thanks{Zhengying Zhu is with the School of Information Science and Technology at ShanghaiTech University,  {\tt\footnotesize zhuzhy1@shanghaitech.edu.cn}}%
\thanks{Xuhao Qin is with the School of Information Science and Technology at ShanghaiTech University,  {\tt\footnotesize qinxh2024@shanghaitech.edu.cn}}%
\thanks{Chenxi Xiao ($^*$ corresponding author) is with the School of Information Science and Technology at ShanghaiTech University,  {\tt\footnotesize xiaochx@shanghaitech.edu.cn}}%
}

\maketitle

\begin{abstract}
Tactile sensors play a crucial role in enabling robots to interact effectively and safely with objects in everyday tasks. In particular, visuotactile sensors have seen increasing usage in two and three-fingered grippers due to their high-quality feedback. However, a significant gap remains in the development of sensors suitable for humanoid robots, especially five-fingered dexterous hands. One reason is because of the challenges in designing and manufacturing sensors that are compact in size. In this paper, we propose HumanFT, a multimodal visuotactile sensor that replicates the shape and functionality of a human fingertip. To bridge the gap between human and robotic tactile sensing, our sensor features real-time force measurements, high-frequency vibration detection, and overtemperature alerts. To achieve this, we developed a suite of fabrication techniques for a new type of elastomer optimized for force propagation and temperature sensing. Besides, our sensor integrates circuits capable of sensing pressure and vibration. These capabilities have been validated through experiments. The proposed design is simple and cost-effective to fabricate. We believe HumanFT can enhance humanoid robots' perception by capturing and interpreting multimodal tactile information.

\end{abstract}

\section{Introduction}

Tactile sensors are critical components in robotic systems, allowing robots to interact efficiently and safely with objects in complex environments. During object manipulation, these sensors provide rich contact information, enabling robots to perceive multimodal data such as pressure, stiffness, texture, and temperature. Such multimodal information has significantly enhanced robotic dexterity, enabling robots to handle fragile objects with appropriate force. This ensures both safety and efficiency in tasks such as organizing kitchen objects \cite{khazatsky2024droid},  assembling items \cite{chen2024general}, and performing robotic surgeries \cite{yu2024orbit}. These advancements have expanded the range of robotic applications.

Recently, visuotactile sensors have attracted considerable attention for their ability to provide high-quality contact information. By integrating cameras into the fingers, these sensors are capable of performing high-resolution surface estimation, and offer a broader range of sensing modalities compared to traditional designs that rely on other techniques \cite{qu2023recent}. Furthermore, since the sensor outputs are camera images, they are compatible with existing machine learning techniques initially developed for computer vision, facilitating seamless integration into robotic learning frameworks originally designed for visual data.

\begin{figure}[t] \centering \vspace{3mm} 
\includegraphics[width=0.98\linewidth]{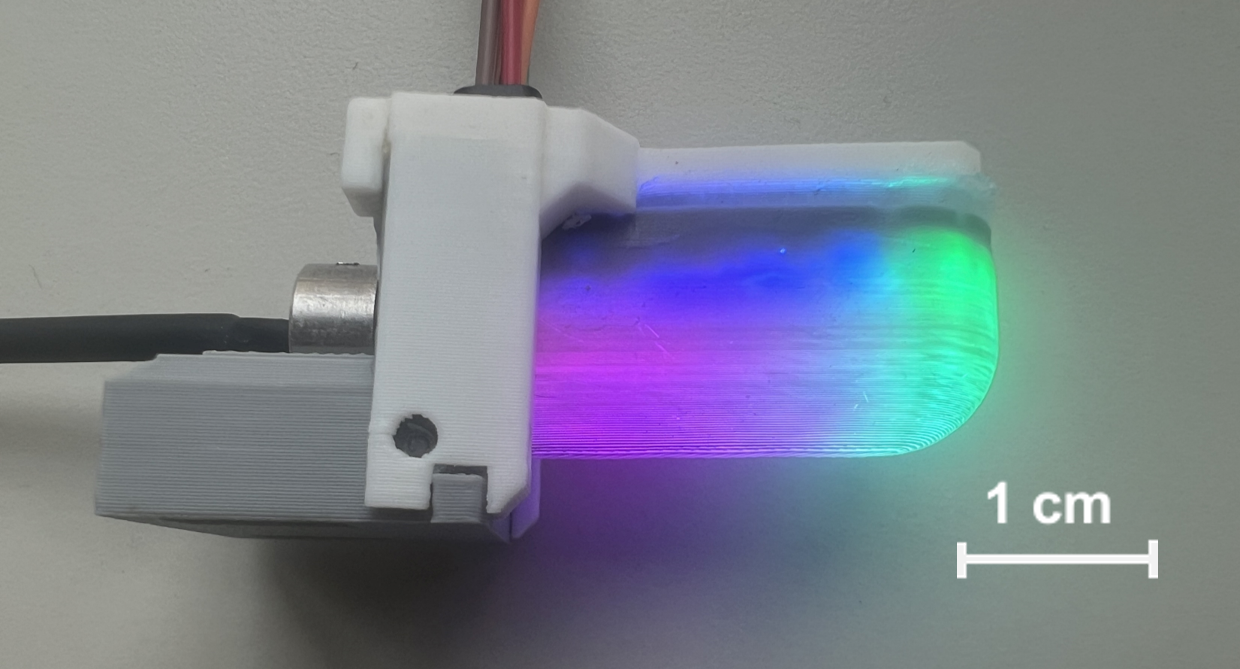} \caption{HumanFT: a visuotactile sensor with a shape similar to a human fingertip. It features multimodal sensing capability due to the integration of sensors within the elastomer.} \label{fig:teaser} 
\end{figure}

As a result of these advantages, various visuotactile sensors have been developed for robotic manipulators, such as the GelSight Wedge \cite{wang2021gelsight}, GelSight Fin Ray sensors \cite{liu2022gelsight, liu2023gelsight}, and three-fingered grippers like the GelSight Svelte \cite{zhao2023gelsight}, to mention a few. However, these manipulators and their installed sensors are relatively bulky. In contrast, the development of tactile sensors for biomimetic anthropomorphic hands remains limited. Anatomical studies show that anthropomorphic fingers are typically small (i.e., around 18-20 mm in diameter \cite{rodriguez2024tissue}), which presents challenges for hardware integration (see Table~\ref{tab:sensor_comparison}).

Human hands also possess more sophisticated multimodal sensing capabilities than those of most current robotic sensors. Existing visuotactile sensors mainly capture contact deformation properties such as 3D shape and stress fields. However, they lack other important sensory modalities present in human skin, such as temperature and vibrotactile sensations \cite{abad2020visuotactile}.  This limits their abilities to provide more informative perception. For instance, vibrotactile feedback can enhance the detection of mechanical frequencies, extending the range from the camera’s typical frame rate of 30-110 Hz to thousands of hertz \cite{li2020review}. Similarly, temperature sensing can help robots avoid hot surfaces, preventing potential damage to the hardware.

To bridge these gaps, we introduce HumanFT, a visuotactile sensor designed for easy integration into robotic anthropomorphic hands. The sensor is compact, with dimensions of $12 \times 20 \times 35$ mm, which resembles a human finger’s distal phalanx. It can directly detect high-frequency vibrations and forces without requiring visual markers, which would otherwise compromise the integrity of the sensor surface during shape reconstruction. From the design perspective, we have simplified the structure to facilitate quick assembly and low-cost fabrication.

In summary, this paper presents the following contributions: 

\begin{itemize}

\item The design of HumanFT, a compact, human finger-like visuotactile sensor.

\item Design and fabrication techniques for a new type of PDMS-based elastomer.

\item A suite of techniques for incorporating force, vibration, and temperature sensing into visuotactile sensors, along with characterization results.

\end{itemize}

\section{Related Works}
\subsection{Tactile Sensors for Robotic Dexterous Hands} 

Over the past few decades, the development of tactile sensors has been driven by the goal of replicating the sensory functions of human skin in robotics and prosthetic limbs \cite{tiwana2012review, kappassov2015tactile}. To this end, numerous tactile sensors have been designed and developed. For instances, the BioTac sensor employs electrical impedance tomography to characterize pressure distributions of fingertip \cite{fishel2012sensing}. Similarly, the uSkin sensor utilizes the Hall effect to detect magnetic fields generated by embedded magnets \cite{tomo2018new,tomo2016design}, to mention a few. Due to their compact size, these sensors have been integrated into various dexterous robotic hands \cite{li2020review}. Examples include Allegro hand that integrates uSkin \cite{tomo2017covering}, DLR Hand which incorporates force/torque sensors \cite{butterfass2001dlr}, and the Shadow Hand, which integrates BioTac sensors.

Recently, visuotactile sensors have attracted attention due to ability to provide enriched sensory data. By embedding cameras within the sensor, these devices can observe surface details at micron-level spatial resolutions, surpassing those of standard robotic cameras. Notable examples include the GelSight series \cite{yuan2017gelsight, zhao2023gelsight, wang2021gelsight,Romero_2020, liu2022gelsight, liu2023gelsight}, Digit \cite{5306070}, GelSlim \cite{taylor2022gelslim}, and the DTact series \cite{lin20239dtact, lin2023dtact}. For a comprehensive review of visuotactile sensors, readers may refer to \cite{abad2020visuotactile, navarro2023visuo, luo2017robotic}.

However, a significant challenge in extending this technology lies in adapting sensor sizes to fit humanoid anthropomorphic fingertips. Unlike two-finger grippers, only very few anthropomorphic hand designs are equipped with multiple high-resolution tactile sensors so far \cite{ford2023tactile}. A comparison of existing sensor designs for this purpose is presented in Table.~\ref{tab:sensor_comparison}. Our proposed solution aims to bridge this gap.

\subsection{Multimodal Visuotactile Sensors}
Human fingertips are capable of sensing various types of information, including force, temperature, and texture (e.g., micro-vibrations). Recent research in robotics has focused on developing multimodal tactile sensors that replicate these capabilities, striving to bridge the gap between human tactile sensing and robotic counterparts \cite{li2020review}. Visuotactile sensors, with their high resolution, are naturally suited for acquiring multimodal information. By incorporating deep learning and visual techniques, these sensors have demonstrated the ability to estimate skin deformation, detect strain fields induced by external forces and torques \cite{yuan2017gelsight}, and identify surface textures thanks to their high resolution \cite{Li_2013_CVPR, hogan2021seeing}. Researchers are also working on integrating temperature sensing \cite{abad2021haptitemp}, multi-axis force measurement \cite{li20233}, and visual feedback for distant sensing \cite{athar2023vistac}. However, integrating all these sensing modalities into a single design remains a challenge. To address this, our proposed solution integrates 3D visual feedback, force measurement, high-frequency vibration sensing, and overtemperature alert into a compact sensor suitable for anthropomorphic robotic hands.

\begin{table}[t]
\centering
\vspace{3mm}
\caption{Comparison of compact visuotactile sensors that resemble human fingers.}
\label{tab:sensor_comparison}
\begin{tabular}{cccc}
\hline\hline
\textbf{Device} & \textbf{Size (mm)} & \textbf{Shape} & \textbf{Multimodal} \\ \hline
GelForce \cite{5306070} & 18 $\times$ 23 $\times$ 36 & 2D Markers & Est. Forces \\ 
TacTip \cite{ford2023tactile, lepora2021towards} & 12 $\times$ 19 $\times$ 25 & 2D Markers & Est. Forces \\ \
Digit & 18 $\times$ 20 $\times$ 27 & 3D & Est. Forces \\ 
GelSight \cite{yuan2017gelsight} & 35 $\times$ 35 $\times$ 60  & 3D & Est. Forces \\ \hline
Ours & 12 $\times$ 20 $\times$ 35 & 3D & \begin{tabular}[c]{@{}l@{}} Measured Forces\\ Vibration\\ Overtemperature\end{tabular} \\ \hline\hline
\end{tabular}
\vspace{-3mm}
\end{table}

\section{Methodology}
\begin{figure*}[t]
    \centering
    \vspace{4mm}
    \includegraphics[width=\linewidth]{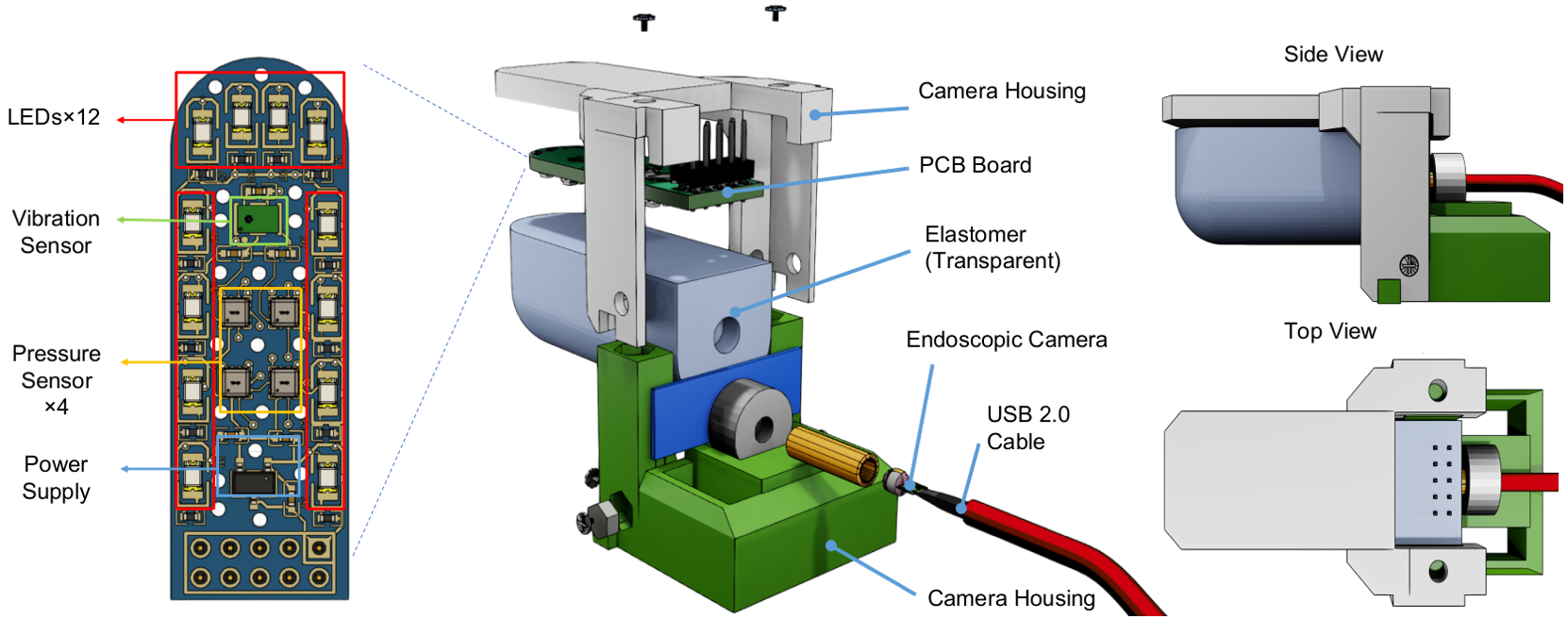}
    
    \caption{\textbf{Exploded view of the visuotactile sensor assembly}. The sensor comprises three main functional components: (1) a transparent elastomer with thermochromic coating, (2) an endoscopic camera, and (3) a PCB equipped with vibration and pressure sensors. This configuration enables multimodal sensing, allowing the detection of elastomer deformation, contact forces, and vibrations.}
    \label{fig:main_structure}
    \vspace{-4mm}
\end{figure*}

\subsection{Sensor Design}
We introduce HumanFT, a compact visuotactile sensor designed to replicate the shape and function of a human fingertip. An exploded view of the sensor is shown in Fig.~\ref{fig:main_structure}. The sensor comprises three main components: a camera, a coated elastomer, and a Printed Circuit Board (PCB). The design and fabrication processes for each component are described below.

\begin{figure}[ht]
    \centering
    \vspace{3mm}
    \includegraphics[width=0.98\linewidth]{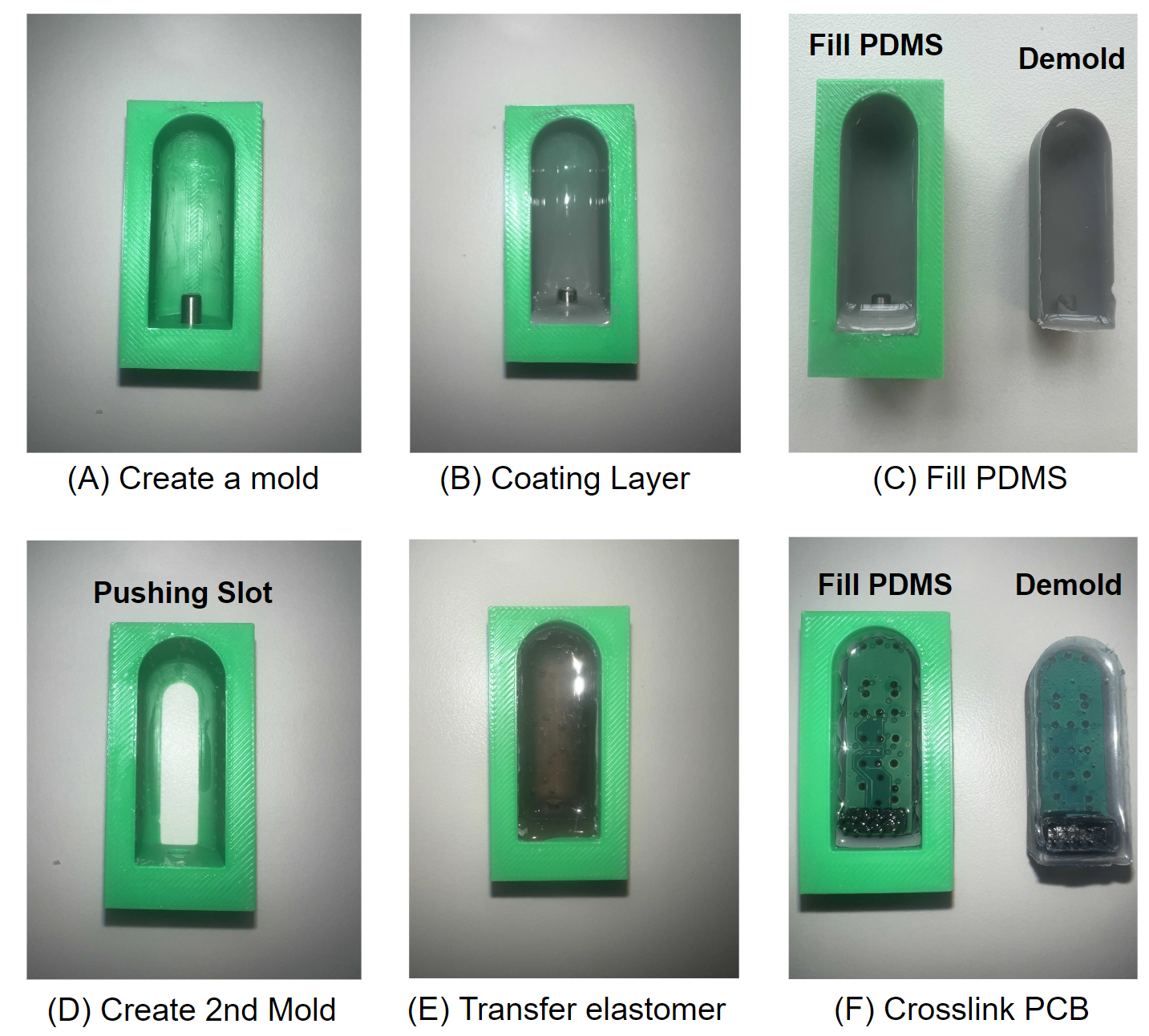}
    \caption{\textbf{Fabrication procedures for the elastomer and sensor assembly}. (A) Create a mold using 3D printing, (B) fabricate the coating layer, (C) fill with PDMS and demold the cured elastomer, (D) create a second mold with a pushing slot, (E) transfer cured elastomer into the second mold, (F) crosslink the PCB by filling with PDMS, and demold by pushing out the sensor.  }
    \label{fig:elastomer}
\end{figure}

\textbf{Elastomer}. The elastomer forms the fingertip's main body and is designed to interact with external objects. We developed a new elastomer that mimics the human finger's distal phalanx. Unlike previous designs featuring a thin elastic layer on a stiff acrylic base \cite{lin2023gelfinger, yuan2017gelsight, ford2023tactile}, our elastomer is a solid, deformable structure. This offers two advantages. First, the thicker elastomer enables the measurement of larger deformations. Second, the solid structure allows contact forces to propagate throughout the fingertip, enabling direct force measurements at the base when additional sensors are placed on the PCB.

The elastomer is fabricated using a mold-casting process. First, a 3D-printed mold (Fig.~\ref{fig:elastomer} (A)) is created from Polylactic Acid (PLA). A small metallic tip is installed to create an opening for the camera that will be placed inside later. Then, the inner surface of the mold is coated with a Lambertian reflective layer to ensure uniform light diffusion and reduce ambient lighting interference (Fig.~\ref{fig:elastomer} (B)). The coating, which is made from Ecoflex 00-31 Silicone Gel (Smooth-On Inc.) mixed with thermochromic pigments, changes color at elevated temperatures, which can be observed via the internal camera. The elastomer coating cures for 1 hour at 70$^\circ$C.

After curing the coating, Polydimethylsiloxane (PDMS) is poured into the mold to form the elastomer's main body (Fig.~\ref{fig:elastomer} (C)). PDMS is known for its high optical transparency, hydrophobicity, and biocompatibility (we used Dow Corning Sylgard 184). The PDMS is left to cure at room temperature for 48 hours. To better replicate the mechanical properties of human muscle tissue, we reduced the hardness of the PDMS from Shore A40 to A18 by incorporating a diluent (DOWSIL DC184 thinner) in a ratio of 11:2.

Next, the elastomer is demolded and transferred to a second mold that includes a pushing slot to simplify the demolding process (Fig.~\ref{fig:elastomer} (D) and (E)). The PCB is placed on top of the elastomer and additional PDMS is used to crosslink the two components. Finally, the crosslinked assembly is removed by pushing the elastomer out from the bottom of the mold (Fig.~\ref{fig:elastomer} (F)).

\textbf{Camera}. The sensor uses an internal camera to visualize the deformations of the elastomer. The camera is positioned to face the fingertip, capturing most of the skin region. We employ the EZ-EN33S-RT endoscope, which integrates the OV9734 sensor with a 140$^\circ$ field of view. This compact camera ($10.9 \times 3.5 \times 3.5$ mm) includes a Ralink RT5350 controller that can directly interface with USB 2.0 bus. Hence, we only need one USB 3.0 hub to provide sufficient bandwidth for connecting cameras for 5 robot fingers, eliminating the need for external hardware like Raspberry Pi arrays. This significantly simplifies sensor integration, improves connection reliability, and reduces costs compared to previous multifingered designs \cite{ford2023tactile, wang2024large}.

\textbf{PCB}. The PCB serves two main functions: (1) providing illumination through LEDs, and (2) enabling multimodal sensing via onboard sensors. For pressure sensing, we integrated four BMP388 pressure sensors (Bosch Inc.). These sensors enable direct measurement of contact forces without the need for visual markers on the elastomer surface, and also reduces the computational overhead associated with visual processing algorithms. Summing the pressure data from all sensors provides an estimate of the normal contact force ($F_z$), while differential pressure measurements allow for the estimation of shear forces ($F_x$, $F_y$). All sensors communicate with the microcontroller via the Serial Peripheral Interface (SPI) bus, and data is transmitted to the computer at a frequency of 125 Hz.

In addition, we integrated a MEMS omnidirectional microphone (ZTS6216) to detect mechanical vibrations. This compact sensor ($2.75 \times 1.85 \times 1$ mm) has a sensitivity of $-38$dBV/Pa, allowing it to detect high-frequency signals in the acoustic range. The sensor's analog output is routed to the microcontroller's ADC for signal processing.

\begin{figure}[t]
    \centering
    \vspace{3mm}
    \includegraphics[width=\linewidth]{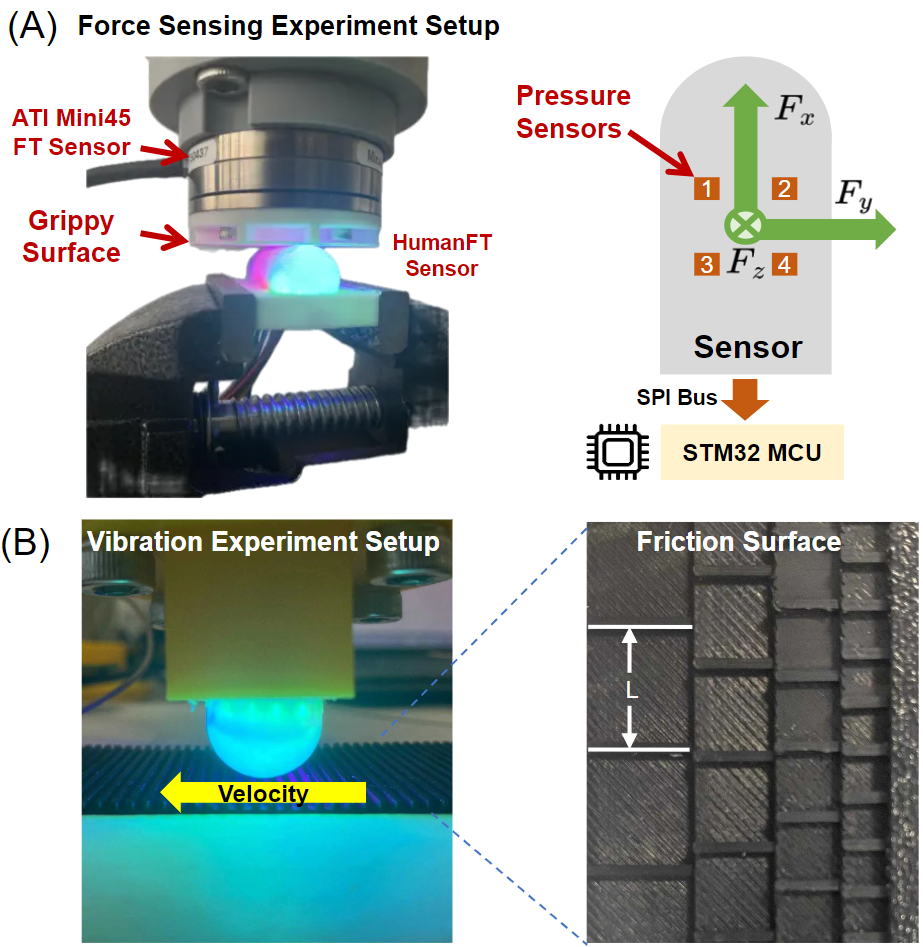}
    \caption{\textbf{Our experimental setups}, for applying (A) external forces, and (B) mechanical vibration to HumanFT sensor.  }
    \label{fig:force_experimental_setting}
    \vspace{-4mm}
\end{figure}

\subsection{3D Reconstruction} \label{sec:met:3d}
In addition to contact force measurement, our sensor captures surface deformations using photometric stereo (i.e., estimating surface normals), followed by Poisson surface reconstruction to obtain the 3D shape of the finger surface. To achieve this, we need to estimate surface normals based on light intensity. A calibration process is required to correlate normal directions with the color intensities captured by the camera. Unlike the method proposed by \cite{yuan2017gelsight}, which does not consider curved surfaces, we adopted the light field model from \cite{gomes2023beyond}. In this model, the light intensity $\bm{I}$ captured by the camera is a function of the surface normals $\bm{N} = [N_x, N_y, N_z]$ and position $\bm{p}=[u, v]$ in pixel space:

\begin{equation} 
\bm{I} = f(N_x, N_y, N_z, u, v) 
\end{equation}

Our goal is to recover the surface normals $\bm{N}$ from the light intensities $\bm{I} = [I_r, I_g, I_b]$:

\begin{equation} 
\bm{N} = g(I_r, I_g, I_b, u, v) 
\end{equation}

\begin{figure*}[t]
    \centering
    \vspace{3mm}
    \includegraphics[width=0.96\linewidth]{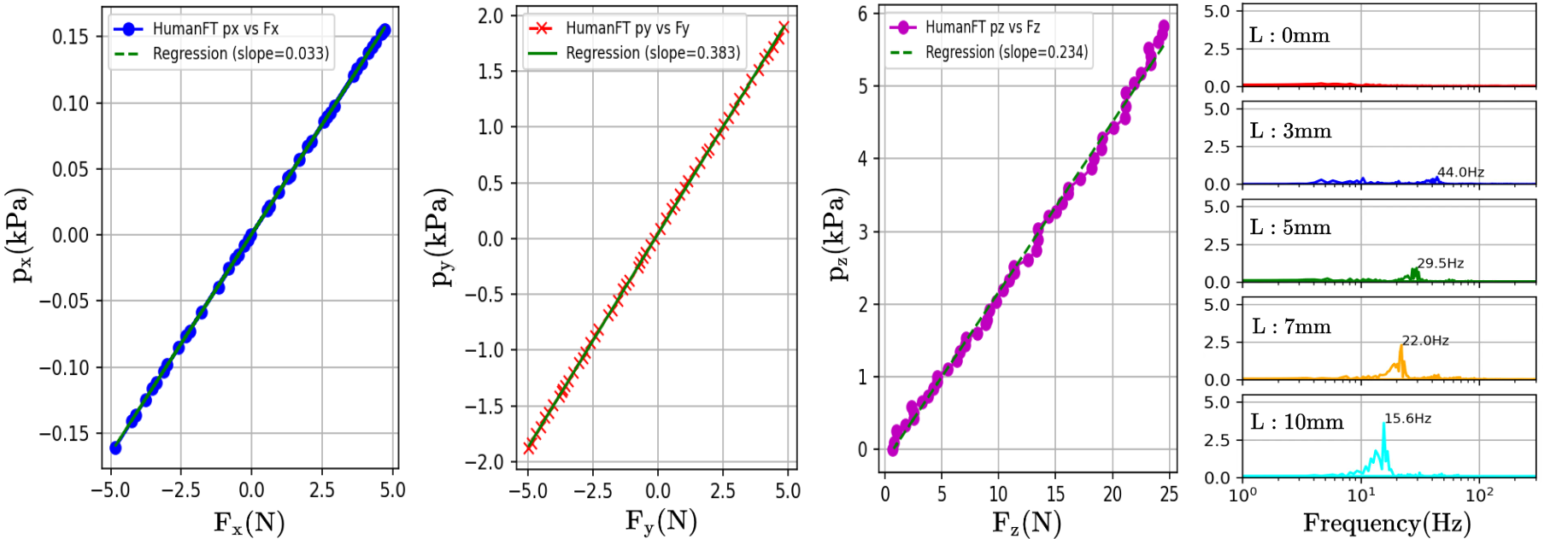}
    \caption{\textbf{Experimental results of force and vibration characterization}. (1) Our sensor's capability to characterize shear forces $F_x$, $F_y$, and normal force $F_z$ using $p_x$, $p_y$, and $p_z$ from the pressure sensors' outputs; (2) sensor's ability to detect mechanical vibrations during sliding motion.}
    \label{fig:characterization_exp} 
    \vspace{-8mm}
\end{figure*}

To achieve this, we developed a pipeline inspired by \cite{gomes2023beyond}. First, we calibrated the camera using a micro (2mm) checkerboard pattern to determine the intrinsic matrix and distortion coefficients outside the elastomer. The calibrated camera then captures images (intensities $I_r, I_g, I_b$) of the elastomer during indentation. Since obtaining ground-truth normals directly in real world is challenging, we created a digital-twin PyBullet simulation to render groundtruth normal maps ($\bm{N}$). Using feature matching, the camera pose error between the real sensor and simulation is minimized. These normal maps, along with the corresponding intensities and positions, are used to train a Multilayer Perceptron (MLP) network for supervised learning.

\section{Experiments}
In this section, we evaluate the performance of the proposed tactile sensor through a series of contact characterization experiments. First, we characterize the sensor's response to external contact, focusing on the pressure measurements from the embedded pressure sensors (Sec.~\ref{exp:force}), as well as vibration detection when sliding on both smooth and rough surfaces (Sec.~\ref{exp:vibration}). Next, we validate the overtemperature detection (Sec.~\ref{exp:temperature}). Then, we demonstrate the capability of our newly designed elastomer to convey contact information through images, enabling 3D reconstruction via photometric stereo (Sec.~\ref{exp:shape}). Finally, we discuss the observations of the experiment and the limitations (Sec.~\ref{exp:discussion}).

\subsection{Force Sensing} \label{exp:force}

{

We first demonstrate that the proposed sensor can measure contact forces as time-series data for $F_x$, $F_y$, and $F_z$. The experimental setup is shown in Fig.~\ref{fig:force_experimental_setting} (A), where an ATI Mini-45 force sensor is mounted on the robot's end effector. A 3D-printed plate is attached to the ATI sensor to increase the surface friction coefficient, preventing slip during applying shear forces. This setup emulates a common scenario in which gripped objects might slip from fingers, generating shear forces $F_x$ and $F_y$ in addition to the normal force $F_z$.

Using the four embedded pressure sensors, we establish a relationship between the sensor readings and the applied external forces. To characterize this relationship, we define:

\begin{equation}
\begin{pmatrix}
p_x \\
p_y \\
p_z
\end{pmatrix}
=
\begin{pmatrix}
\frac{1}{2} & \frac{1}{2} & -\frac{1}{2}  & -\frac{1}{2}  \\
-\frac{1}{2}  & \frac{1}{2}  & -\frac{1}{2}  & \frac{1}{2}  \\
\frac{1}{4} & \frac{1}{4} & \frac{1}{4} & \frac{1}{4}
\end{pmatrix}
\begin{pmatrix}
p_1 \\
p_2 \\
p_3 \\
p_4
\end{pmatrix},
\end{equation} where $p_x$, $p_y$, and $p_z$ are the rectified sensor outputs that reflect external forces, given the raw pressure sensor readings $p_1$, $p_2$, $p_3$, and $p_4$ (sensor indexes are labeled in Fig.~\ref{fig:force_experimental_setting} (A)).

We then independently characterize the response of $p_x$, $p_y$, and $p_z$ to the external forces $F_x$, $F_y$, and $F_z$. To obtain these force measurements, we use the ATI Mini-45 sensor to provide accurate readings of $F_x$, $F_y$, $F_z$. We visualize the relationships between $p_x$, $p_y$, $p_z$ and the corresponding forces $F_x$, $F_y$, $F_z$, as shown in Fig.~\ref{fig:characterization_exp}.

First, we characterize the sensor's response to the shear forces $F_x$ and $F_y$. The shear force is generated through static friction by fixing $F_z$ at 20N to prevent slipping, and then controlling the robot’s end effector to move in the $+x$, $-x$, and $+y$, $-y$ directions. By recording both the sensor’s response and the Force-Torque (FT) sensor readings, we observe a linear relationship between $p_x$ and $F_x$, as well as between $p_y$ and $F_y$. The sensitivity of $p_x$ to $F_x$ is 0.033 kPa/N, while the sensitivity of $p_y$ to $F_y$ is 0.383 kPa/N. Note that the sensitivity differs along the two shear force directions, which we attribute to the asymmetrical shape of the elastomer.

Similarly, we characterize the sensor's response $p_z$ to the external normal force $F_z$. A linear regression of the data reveals a linear relationship between $p_z$ and $F_z$, with a sensitivity of 0.234 kPa/N.
This indicates that our sensor can characterize both shear forces and the normal force.

\vspace{-1mm}

\subsection{Vibration Sensing} \label{exp:vibration}
Next, we validate the sensor’s ability to characterize vibration signals by sliding it over surfaces with varying roughness. We slide our sensor over a flexible surface (3D-printed using thermoplastic polyurethane, TPU) featuring grating ridges with intervals $L$ of 0, 3, 5, 7, and 10 mm, respectively. By sliding the sensor over these surfaces, we record the microphone’s output signal and then transform the data into the frequency domain. While the microphone sensor covers the acoustic frequency range, we found that most mechanical vibrations occurred below 1 kHz, aligning with the human skin’s detection range \cite{oroszi2020vibration, 10.7554/eLife.46510}. We used a Fast Fourier Transform (FFT) with 44,277 points at a sampling frequency of 25,641 Hz, resulting in a frequency resolution of approximately 0.5791 Hz.

The transformed results are shown in Fig.~\ref{fig:characterization_exp}. From the results, we observe that for $L = 0$ mm (a smooth surface), the signal peak is not prominent. Signal peaks are evident for all other $L$ values (non-smooth surfaces). For larger distances $L$ between ridges, the peak frequency decreases, but the vibration amplitude increases. In addition, the sensor can detect harmonic vibration waves beyond the camera’s frame rate, which increases the dynamic response range that sensor can characterize.

\subsection{Overtemperature Detection} \label{exp:temperature}

We demonstrate the sensor’s overtemperature detection capability using thermochromic paint coating. To validate this feature, the elastomer was heated to 65$^\circ$C on a hot plate. A comparison of the surface color before and after heating is shown in Fig.~\ref{fig:temperature}. As observed through the embedded camera, the heated region changes color to yellow (bright region at the bottom), which is clearly distinguishable. The color change occurs in approximately one second, indicating immediate overtemperature alert can be achieved.

\begin{figure}
    \centering
    \includegraphics[width=0.95\linewidth]{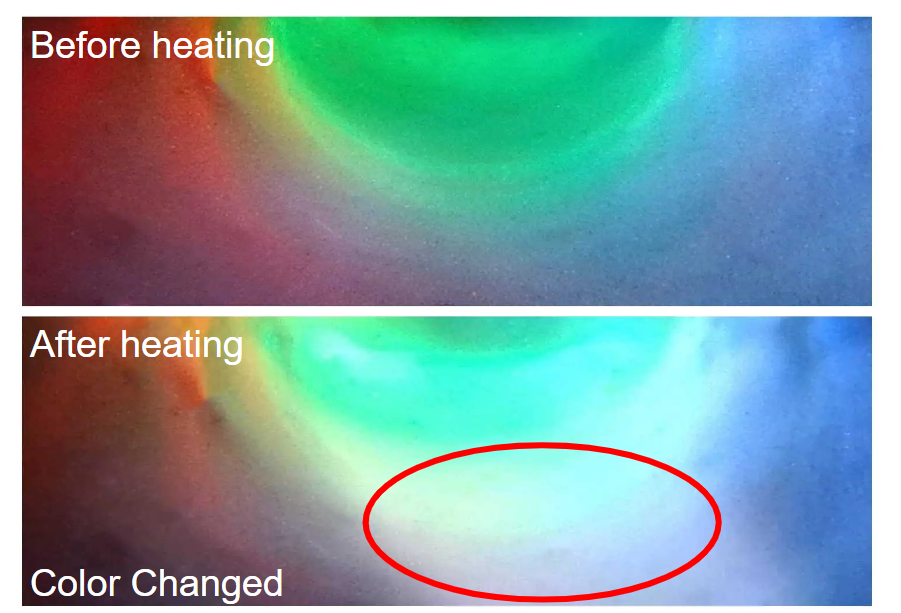}
    \caption{\textbf{Surface color changes observed after heating the sensor on a hot plate}. This demonstrates the overtemperature detection capability.}
    \label{fig:temperature}
    \vspace{-6mm}
\end{figure}

\subsection{Shape Characterization} \label{exp:shape}
\begin{figure}[ht]
    \centering
   
    \includegraphics[width=0.98\linewidth]{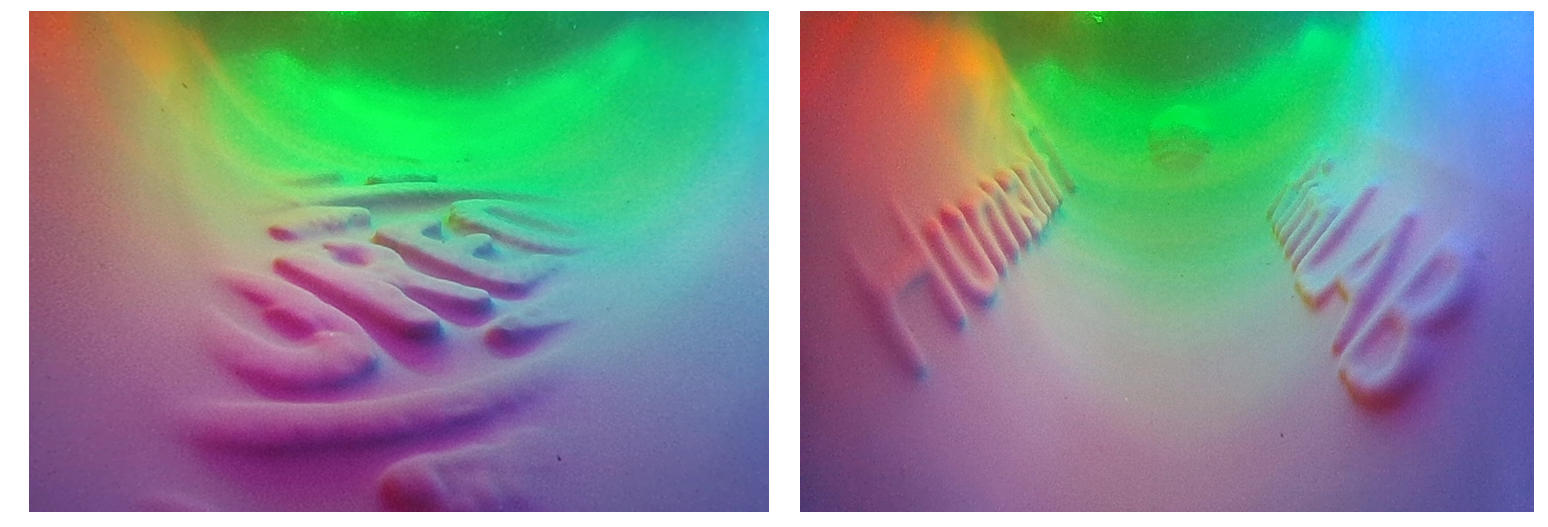}
    \caption{\textbf{Camera views from inside of elastomer}. Both the bottom surface and two side surfaces can be used to detect fine-grained surface details. }
    \label{fig:camera_inside_texts}
\end{figure}

\begin{figure}[ht]
    \centering
    \vspace{2mm}
    \includegraphics[width=\linewidth]{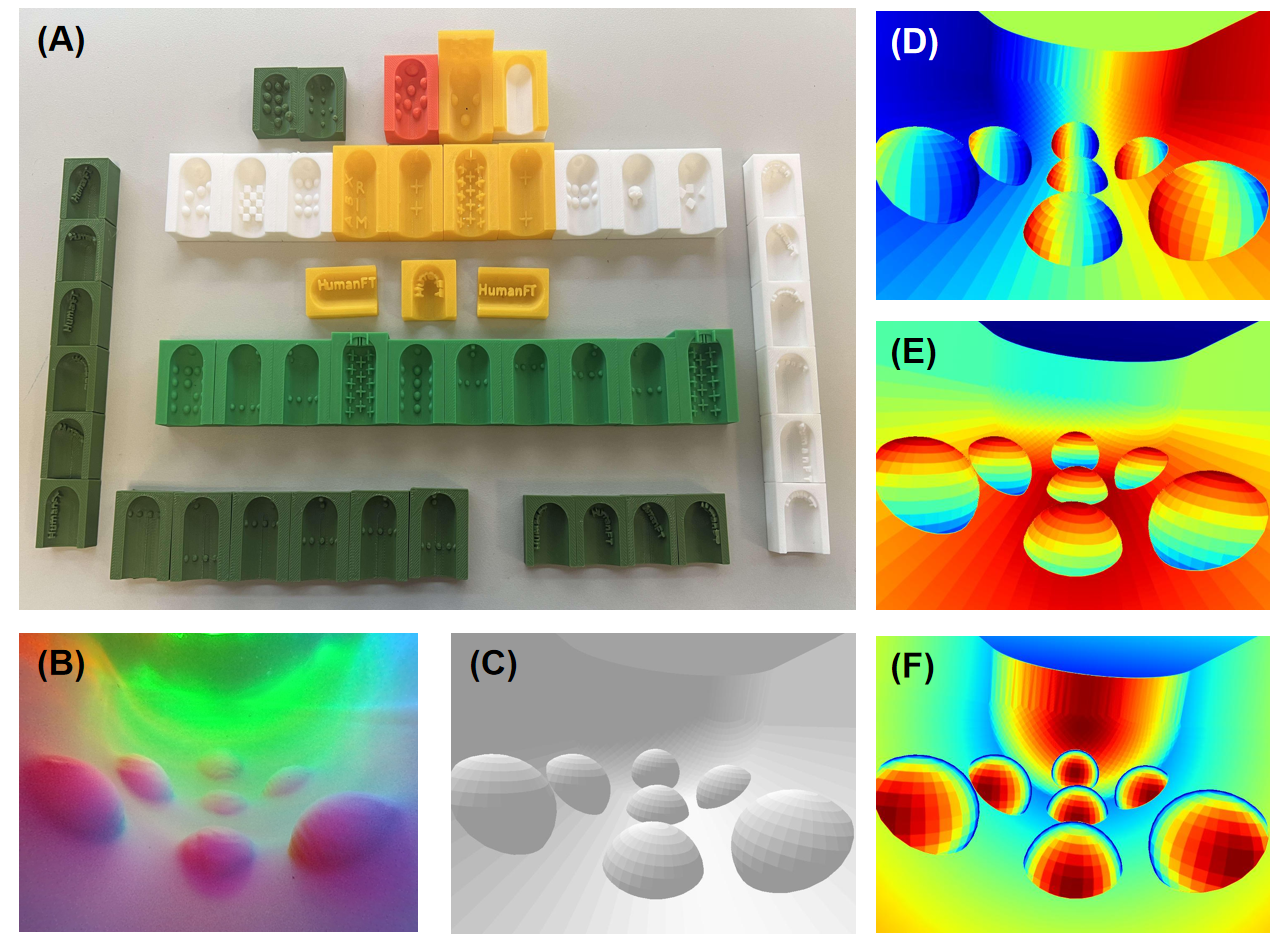}
    \caption{\textbf{Acquisition of normal groundtruth}. (A) Indenter molds used for sensor calibration, (B) sensor observation during indentation. (C) depth map from simulation. (D-F) normals obtained from simulation rendering.  }
    \label{fig:calibration}
    \vspace{-5mm}
\end{figure}

Our sensor can provide high visual quality (Fig.~\ref{fig:camera_inside_texts}). Based on this, 3D reconstruction was performed using the method described in Sec.~\ref{sec:met:3d}. To accomplish this, a neural network was trained to estimate normal vectors from RGB images. This was achieved by capturing a dataset of real images, where a series of 3D-printed indenter molds were used to press onto the fingertip, while concurrently using simulations to collect ground-truth normal vectors. This procedure is illustrated in Fig.~\ref{fig:calibration}.

\begin{figure}[ht]
    \centering
    \includegraphics[width=\linewidth]{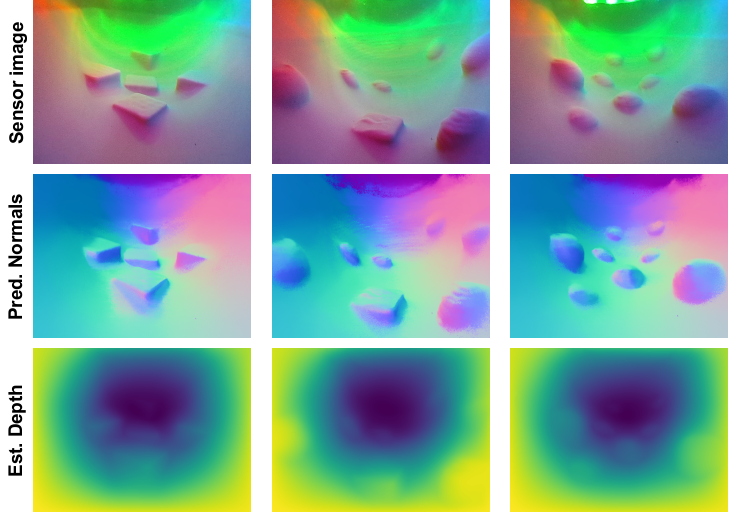}
    \caption{\textbf{Results of 3D reconstruction}. This includes: original image, estimated normals, and estimated depth using Poisson surface reconstruction.  }
    \label{fig:poisson}
    \vspace{-1mm}
\end{figure}

The trained neural network successfully predicted surface normals. We visualize the reconstruction results for three indenter molds in Fig.~\ref{fig:poisson}, where predicted normals (colors indicating different directions) and the estimated depth maps are displayed. From the depth maps, we observe the locations of indentation accurately correspond to the sensor images, demonstrating the effectiveness of reconstruction.

\subsection{Discussions} \label{exp:discussion}
Through our experiments, we have demonstrated that the proposed visuotactile sensor effectively detects contact and provides multimodal sensing. Compared to previous visuotactile sensors, our sensor not only performs surface reconstruction but is also capable of capturing multimodal information. The key contribution that enables these new capabilities is the use of a solid yet compliant elastomer structure, which not only offers high optical transparency but also provides force propagation paths to the sensors underneath. To our knowledge, this is the first visuotactile sensor equipped with concurrent sensing capabilities for visuotactile feedback, forces, vibrations, and temperature.

The sensor can be manufactured at a low cost (around 50 USD). The main expense is the camera, which costs around 40 USD. The PCB costs approximately 10 USD, while the other components incur negligible costs.

One limitation is pressure sensor's reading accuracy might be affected by temperature rise caused by LEDs during system startup (only lasts for a few minutes). Another limitation of our sensor is that its 3D reconstruction performance is not as accurate as that of the GelSight sensor of flat surfaces. We believe this is due to sensor's calibration. The use of a completely soft elastomer may propagate strain to broader regions and, in some cases, slightly shift the camera's position due to elastic effects under very large contact forces. As a result, we occasionally observed local misalignment between the ground-truth normals and the indented regions. One technique that might help address this issue is to incorporate soft-body simulation to improve the accuracy of ground-truth normal alignment \cite{si2024difftactilephysicsbaseddifferentiabletactile, gomes2023beyond}, and develop a better camera housing structure.

\section{Conclusions}
In this paper, we developed HumanFT, a human-like fingertip tactile sensor with dimensions of $12 \times 20 \times 35$ mm. The sensor has a shape that is more compact than most previous designs and highly resembles a human fingertip. The proposed sensor is also low-cost, and simple to fabricate. By utilizing a new elastomer design based on a PDMS solid body and Ecoflex 00-31 soft coating layer, we created a direct force propagation path to the bottom of the sensor. This design enables the direct sensing of multimodal signals using onboard sensors, complementing visual feedback for the acquisition of forces and high-frequency vibrations beyond the camera's frame rate. Furthermore, the coating layer incorporates thermochromic paint, which can provide overtemperature alerts when touching hot surfaces.

Future work will focus on improving 3D reconstruction performance and demonstrating the sensor's functionality across multiple fingertips on dexterous robotic hands. We believe that this sensor can bridge the gap in collecting and analyzing multimodal data, which holds promise for acquiring more advanced datasets for robotic manipulation tasks that require multimodal sensing.

\section{Acknowledgement}
We thank Zengfan Xing and for exploring gel casting, Ziyuan Tang for the creation of molds, Pei Lin for discussions on reconstruction algorithms, and Xiyan Huang for assistance in data collection. \textbf{We plan to release CAD files and circuit design upon paper acceptance.}

\clearpage

\bibliography{egbib}
\bibliographystyle{IEEEtran}
\end{document}